\documentclass[twoside]{article}

\usepackage[accepted]{aistats2023}
\usepackage[round, authoryear]{natbib}

\usepackage{graphicx} % for pdf, bitmapped graphics files
\usepackage{amsmath, amsthm} % assumes amsmath package installed
\usepackage{amssymb}  % assumes amsmath package installed
\usepackage{tikz}
\usepackage{algorithm,algorithmic}
\usepackage{array}

\usepackage{url}
\usepackage[pdfstartview=FitV, urlcolor=blue]{hyperref}

\newcommand{\bv}{\boldsymbol{v}}
\newcommand{\bx}{\boldsymbol{x}}
\newcommand{\by}{\boldsymbol{y}}
\newcommand{\bz}{\boldsymbol{z}}
\newcommand{\ba}{\boldsymbol{a}}
\newcommand{\x}{\bx}
\newcommand{\y}{\by}
\newcommand{\z}{\bz}
\newcommand{\bdy}{\partial \Omega}
\newcommand{\src}{\bx_0}
\newcommand{\J}{\mathcal{J}}
\newcommand{\pathset}{\mathcal{P}}
\newcommand{\expect}{\mathbb{E}}
\newcommand{\Ktilde}{\tilde{K}}
\newcommand{\Kcell}{K_{\text{cell}}}
\newcommand{\Ktildecell}{\tilde{K}_{\text{cell}}}
\newcommand{\sigtilde}{\tilde{\sigma}}
\newcommand{\Khat}{\hat{K}}
\newcommand{\prob}{\mathbb{P}}
\newcommand{\R}{\mathbb{R}}
\newcommand{\Gcell}{\mathcal{G}}
\newcommand{\Gcaught}{\mathcal{G}_c}
\newcommand{\Gtime}{\mathcal{G}_t}
\newcommand{\Gvisit}{\mathcal{G}_n}
\newcommand{\Gobserve}{\mathcal{G}_{\text{ob}}}
\newcommand{\Xobserve}{X_{\text{ob}}}
\newcommand{\Gpredict}{\mathcal{G}_{\text{pd}}}

\DeclareMathOperator*{\argmin}{arg\,min}

\newtheorem{theorem}{Theorem}
\newtheorem{lemma}{\textbf{Lemma}}

\begin{document}

% If your paper is accepted and the title of your paper is very long,
% the style will print as headings an error message. Use the following
% command to supply a shorter title of your paper so that it can be
% used as headings.
%
%\runningtitle{I use this title instead because the last one was very long}

% If your paper is accepted and the number of authors is large, the
% style will print as headings an error message. Use the following
% command to supply a shorter version of the authors names so that
% they can be used as headings (for example, use only the surnames)
%
%\runningauthor{Surname 1, Surname 2, Surname 3, ...., Surname n}

\twocolumn[

\aistatstitle{ Surveillance Evasion Through Bayesian Reinforcement Learning } 

\aistatsauthor{ Dongping Qi \And David Bindel \And  Alexander Vladimirsky }

\aistatsaddress{ Cornell University \And  Cornell University \And Cornell University } 
]

\begin{abstract}
We consider a task of surveillance-evading path-planning in a continuous setting.
An Evader strives to escape from a 2D domain while minimizing the risk of detection (and immediate capture).
The probability of detection is path-dependent and determined by the spatially inhomogeneous surveillance intensity, which is fixed but a priori unknown and gradually 
learned in the multi-episodic setting.  
We introduce a Bayesian reinforcement learning algorithm 
that relies on a Gaussian Process regression (to model the surveillance intensity function based on the information from prior episodes),
numerical methods for Hamilton-Jacobi PDEs (to plan the best continuous trajectories based on the current model), and 
Confidence Bounds (to balance the exploration vs exploitation).
We use numerical experiments and regret metrics to highlight the significant advantages of our approach compared to traditional graph-based algorithms of reinforcement learning. 
\end{abstract}

%%%%%%%%%%%%%%%%%%%%%%%%%%%%%%%%%%%%%%%%%%%%%%%%%%%%%%%%%%%%%%%%%%%%%%%%%%%%%%%%
\section{INTRODUCTION}
\label{s:intro}
% !TEX root = root.tex

Path planning is a standard task in robotics, but it  becomes much harder if we need to account for stochastic perturbations, adversarial interactions, and incomplete information about the dynamics or the environment.
With repeated tasks, Reinforcement Learning (RL) provides a popular framework for optimizing the system performance based on 
the information accumulated in prior episodes while ensuring the asymptotic convergence to the globally optimal solution as the number of 
planning episodes grows \citep{sutton2018reinforcement}.  
In continuous setting, most applications of RL are focused on learning the process dynamics \citep{recht2019tour} using frequent  
or continuous observations of the system state.  Our focus here is on a rather different class of problems, where the controlled dynamics are known, 
but the process termination is a random event, whose 
probability distribution is not only trajectory-dependent but also a priori unknown.

More specifically, we study the online path planning strategy for an Evader (E), who attempts to escape a region while minimizing the probability of being detected en route.
Surveillance intensity imposed by the opponent(s)  is assumed to be spatially inhomogeneous, making the choice of E's capture-evading trajectory important.
This general setting
 is motivated by  
 prior work on 
 environmental crime modeling \citep{cartee2020control} and surveillance avoidance \citep{gilles2020evasive, cartee2019time}.    
However, unlike in those prior papers, here the surveillance intensity is 
initially unknown
to E, who needs to learn it on relevant parts of the domain through multiple planning episodes\footnote{
We note that our setting is also quite different from the classical {\em Surveillance-Evasion Games} \citep{dobbie1966solution, LewinOlsder, takei2014efficient}, in which the surveillance intensity changes dynamically through adversarial motion of the Observer (O), leading to a differential zero-sum game between E and O, who have immediate and full information of the opponent's actions.}.
Whenever E is spotted, this results in capture and immediately terminates the trajectory.  But the evidence obtained about the surveillance intensity on already traced parts of that trajectory can be used to improve the planning in future episodes.
Such multi-episodic setting might seem unusual in capture/surveillance avoidance, but it arises naturally in several applied contexts including the environmental crime modeling.  E.g., in many parts of Brazil, illegal forrest loggers are primarily subsistence farmers in need of firewood for family use \citep{chen2021modeling}. When apprehended, their punishment is usually wood confiscation plus sometimes a small fine. Repeat offenders are common, and they also share information with each other on paths taken or locations where they were caught in the past.  Another example comes from asymmetries in modern warfare, where many types of UAVs become increasingly cheap -- particularly compared to effective air defense systems for large geographic areas.

We develop an approach that balances the exploration against exploitation and uses spatial correlations for efficient learning.
Our algorithm relies on numerical methods for solving PDEs \citep{sethian1996fast}, statistical estimates with censored data \citep{shorack2009empirical}, Gaussian process (GP) regression \citep{williams2006gaussian}, and strategic exploration techniques from RL \citep{kocsis2006bandit, azar2017minimax}.
To simplify the exposition, our method is described and benchmarked here under the assumption that the Evader is {\em isotropic} (i.e., E can change the direction of motion instantaneously and the available speed depends on its current position, but not on its chosen direction).  However, our main ideas are more broadly applicable, and the approach is also suitable for more realistic (anisotropic) agent dynamics.

We start by reviewing the basic problem with {\em known} random termination (capture) intensity in section \ref{s:known_K}. 
Section \ref{s:unknown_K} poses the problem with {\em unknown} intensity and defines performance metrics for episodic path planning.
We follow this with a review of algorithms for 
strategic exploration on graphs developed for finite horizon Markov Decision Processes 
and explain why their usefulness is rather limited in our continuous setting.
In section \ref{s:ourAlgs} we describe a new approach based on Bayesian models of surveillance intensity function
and episodic path planning based on %those model modified with 
the Confidence Bounds.  
We show that piecewise-continuous models lead to simpler algorithms but are usually outperformed by models based on GP-regression.
The advantages of our methods are illustrated on several sample problems in section \ref{s:experiments}.
We conclude by considering possible future extensions in section \ref{s:conclusions}.

\section{PATH PLANNING WITH KNOWN INTENSITY}
\label{s:known_K}
% !TEX root = root.tex

Suppose E starts at $\bx$ in some compact domain $\Omega \subset \R^2$ and moves with isotropic speed $f(\bx)$. 
The motion of E is governed by:
\begin{equation}\label{eq:dynamics}
 \by'(s) = f\bigl(\by(s)\bigr)\ba(s), \quad \by(0) = \bx,
\end{equation}
where $\ba: \R \to S^1$ is a measurable control function specifying the direction of motion at every moment.

Define $T_{\ba} = \min \{s\ge 0 ~ | ~ \by(s) \in \bdy \}$ as the domain-exit time 
if E starts from $\x$ and uses the control $\ba(\cdot)$.
The  
location-dependent surveillance intensity is
a smooth, positive function $K(\bx),$
which in this section is assumed to be fully known in advance.
If E decides to follow a trajectory $\by(\cdot)$, the probability of remaining undetected until time $t$ is
\begin{equation}
\label{eq:capture_prob}
\prob(S\leq t) = 1 - \exp\left( -\int_0^t K(\by(s))ds \right),
\end{equation}
where $S$ is the random time when E is spotted and immediately captured, thus terminating the trajectory.
E's goal is to maximize its probability of reaching $\bdy$
 or, equivalently, to minimize the cumulative intensity:
\begin{equation}\label{eq:J} 
 \J(\bx,\ba(\cdot)) = \int_0^{T_{\ba}} K(\by(s))ds.
\end{equation}
As usual in dynamic programming,
the \textit{value function} $u(\bx)$ is defined to encode the result of
optimal choices
\begin{equation}\label{eq:valfunc}
  u(\bx) = \inf_{\ba(\cdot)} \J(\bx,\ba(\cdot)),
\end{equation}
and can be found as a solution of a Hamilton-Jacobi-Bellman equation \citep{bardi2008optimal}.
Here we focus on isotropic dynamics/intensity; i.e., $f$ and $K$ do not depend on $\ba$, which further simplifies the PDE to the following Eikonal equation:
\begin{equation}\label{eq:eikonal}
  \begin{split}
    \left|\nabla u(\bx) \right|f(\bx) & = K(\bx); \\
          u\left(\bx\right) & = 0, \quad \forall \bx\in\bdy. 
  \end{split}
\end{equation}
This isotropic setting creates a one-to-one correspondence between a control function and a path.
From now on, we interchangeably use the terms ``determining a control function'' and ``selecting a path''.

In general, \eqref{eq:eikonal} often does not have a classical solution, but always has a unique Lipschitz continuous \textit{viscosity solution} \citep{bardi2008optimal}.
Wherever $\nabla u$ exists, the optimal $\ba$ is opposite to the gradient direction ($\ba_* = -\nabla u /|\nabla u|$).
The set on which $u$ is not differentiable has measure zero and is comprised of all starting positions from which the optimal trajectory to $\bdy$ is not unique.

Efficient numerical methods for solving \eqref{eq:eikonal} have been extensively studied in the last 25 years.
Many of these algorithms take advantage of the \textit{causality} found in upwind finite-difference discretizations: a gridpoint only depends on its smaller adjacent neighbors, making it possible to solve the system of discretized equations non-iteratively.
We choose Fast Marching Method(FMM) \citep{sethian1996fast}, which is a Dijkstra-like 
algorithm that has $O(N \log N)$ computational complexity when solving \eqref{eq:eikonal} on a grid with $N$ gridpoints.
Once the value function is approximated, an optimal trajectory can be obtained starting from any $\x \in \Omega$ by gradient decent in $u$  until reaching $\bdy$.

\section{MULTI-EPISODIC PLANNING WITH UNKNOWN $K(\bx)$}
\label{s:unknown_K}
% !TEX root = root.tex

With a known $K(\bx)$, E should persist in choosing a fixed optimal path deterministically, even though the outcomes (whether and where E is captured) may be different every time.
But what if $K(\x)$ is a priori unknown and only learned gradually by trying different paths?
If E faces a repeated task of surveillance-avoidance en route to $\bdy$,
a natural interpretation is to find a path-selection {\em policy} that 
optimizes some long-term performance metric.
This is a
reinforcement learning (RL) problem, with information gradually collected in a sequence of episodes.
Clearly, if E selects enough random trajectories that sufficiently cover the whole region, eventually $K(\bx)$ can be approximately recovered.
However, this approach is inefficient since E's goal is to learn $K(\bx)$ only on those parts of $\Omega$ that are relevant to reduce the frequency of captures,
asymptotically approaching the probability along the truly optimal trajectory, which would be chosen if $K(\x)$ were known.

To assess the long-term performance of a policy, a frequently considered criterion is \textit{regret}, which is 
expected excess of cost due to not selecting the optimal control for all episodes.
Letting $\Delta_i$ be the indicator of whether E is captured during the $i$th episode,
we define the experimentally observed \emph{excess rate of captures}
\begin{equation}\label{eq:regret_measures_2}
 \mathfrak{S}_j = \frac{1}{j}\sum_{i=1}^j \left( \Delta_i - W_* \right), 
\end{equation}
where $W_* = 1 - \exp( - u(\src))$ is the minimum capture probability.
If an optimal $\ba_*(\cdot)$ were used in each episode,
the regret  
$\mathfrak{S}_j$ would converge to $0$ as $j \rightarrow \infty.$

\subsection{Prior work: RL algorithms on graphs}
\label{s:discrete_ucb_algorithm}
% !TEX root = root.tex

Before delving into the continuous problem, we first present a discrete version
on a finite directed graph $G$ and examine the possibility of applying well-known RL algorithms. 
Suppose E starts from a node $\bv_0$, moves between adjacent nodes, and tries to avoid capture en route to a set of target nodes $\Xi$.
We assume that a transition along any edge $e$ incurs some capture probability\footnote{ 
If the graph is embedded in a continuous domain $\Omega$, this $\Psi_e$ can be computed from \eqref{eq:capture_prob}, provided $t$ is the time needed to traverse that edge $e$, which is parametrized by a path $\by(s).$}
$\Psi_e \in (0,1).$ 
If $\pathset$ denotes the set of all paths from $v_0$ to $\Xi$,
we would prefer to use $p\in\pathset$ which maximizes the probability of not being captured up to $\Xi$; i.e., $\max_{p\in\pathset} \prod_{e\in p} (1 - \Psi_e)$ or, equivalently, $\min_{p\in\pathset} \sum_{e\in p} -\log(1 - \Psi_e)$.
When all $\Psi_e$'s are known, this becomes a standard shortest path problem, with $C_e = -\log(1 - \Psi_e)$ interpreted as edge weights, and the classical Dijkstra's method can
solve it efficiently.  
Alternatively, this can be viewed as a simple Markov Decision Process (MDP) by adding an absorbing ``captured state'' node $\bv_c$.
In this MDP interpretation, an action corresponds to a choice of the next attempted edge $e$, a capture event is modeled as a transition to $\bv_c$, and a unit reward is earned only upon reaching $\Xi$.

Of course, we are interested in the case where $\Psi_e$s are {\em not} known in advance, and it would seem natural to address this by any of the RL algorithms  developed to maximize 
the expected return in MDPs with unknown transition functions.  This includes the $Q$-learning \citep{watkins1992q}, temporal difference methods \citep{sutton1988learning}, Thompson sampling \citep{thompson1933likelihood}, and the techniques based on \textit{upper confidence bounds} (UCB) 
\citep{auer2002using}.
The latter served as a basis for a popular model-free Upper Confidence bounds on Trees (UCT) algorithm \citep{kocsis2006bandit},
in which the evidence gathered in previous episodes is used to estimate the rewards of all (state, action) pairs, but 
the exploration of less visited pairs is encouraged by adding a bonus term 
proportional to each estimate's standard deviation.
The same idea is also used in more recent model-based methods \citep{dann2015sample, azar2017minimax}, in which prior evidence is used to model 
the transition probabilities and the value function is computed in each episode based on the current model but with similar bonus terms added to encourage the exploration.

While the above algorithms were originally designed for MDPs with a fixed finite horizon, they can also be adapted to our ``exit time'' case. 
For simplicity, suppose that $\Omega$ is discretized using a uniform Cartesian grid of nodes $V,$
with each interior node $\bv$ connected by edges to its 8 closest neighbors -- corresponding to 8  
actions (directions of motion) available at that node.  We will use $\mathcal{E}(\bv)$ to denote all edges available at $\bv$ and 
$\mathcal{E}$ for the set of all edges in the graph.
Using the MDP interpretation,
the process terminates upon reaching $\Xi \subset V$ (which discretizes $\bdy$) or $\bv_c$ (in case of capture).
To implement UCT, we maintain the statistics $N_{\bv}$ (and $N_{e}$) on how many times each state (and each edge -- or state-action pair)
is visited over multiple episodes, with $Q_e$ encoding the fraction of those visits on which 
E traversed $e$ but was captured before reaching $\Xi$.
The selection of nodes is summarized in Algorithm \ref{alg:UCT}, with the parameter $\lambda>0$ regulating the rate of exploration.  
\begin{algorithm}
  \caption{UCT: model-free planning on a graph}
    \begin{algorithmic}\label{alg:UCT}
    \renewcommand{\algorithmicrequire}{\textbf{Input:}}
    \renewcommand{\algorithmicensure}{\textbf{Output:}}
    \STATE Set $Q_e = 0, \; N_{e} = 0, \; N_{\bv} = 0$ for all $\bv$ and $e$ 
      \WHILE {$t = 1:T$}
	 \STATE capture = \texttt{search}$(\bv_0);$
      \ENDWHILE
      
      \vspace*{2mm}
      
      \hrule
      
      \vspace*{2mm}

      \STATE 
      \hspace*{-4mm}Function \texttt{search} :\\
      \REQUIRE  current node $\bv \not \in \Xi$\\
      \ENSURE capture flag $c$\\
      	 \STATE $N_{\bv} = N_{\bv} + 1;$ 
	 \STATE $\hat{e} = \argmin_{e \in \mathcal{E}(\bv)} Q_e - \lambda \sqrt{\log (N_{\bv})/ \max \left(N_{e}, 1 \right)};$
      	 \STATE $N_{\hat{e}} = N_{\hat{e}} + 1;$ 
	 \STATE $\hat{\bv} = \text{\texttt{attempt\_transition}}(\bv, \hat{e});$
	 \IF {$\hat{\bv} == \bv_c$}
	 	\STATE 
		$c = 1;$
	 \ELSIF {$\hat{\bv} \in \Xi$} 
	 	\STATE 
		$c = 0;$
	\ELSE
		\STATE $c = \text{\texttt{search}}(\hat{\bv});$
	\ENDIF
		\STATE $Q_{\hat{e}} = 
		   \left[ (N_{\hat{e}}-1)Q_{\hat{e}} \, + \, c \right] / N_{\hat{e}};$  
  \end{algorithmic} 
\end{algorithm}

The model-based version on a graph (which we will denote Alg-D, for ``discrete'') is implemented on the same grid, but relies on learning $\Psi_e$ for the relevant edges.
We maintain statistics $N_e$ (and $\phi_e$) on how many times a visit (and capture) happen for each edge $e$.
An estimate of $\Psi_e$ is computed as $\tilde{\Psi}_e = \phi_e \big/ N_e$ and the confidence-bound-modified version is 
\begin{equation}
\label{eq:alg_D_ucb}  
\hat{\Psi}_e = \max\left\{\underline{\Psi},\, \tilde{\Psi}_e - \sqrt{\frac{\log(T |\mathcal{E}| /\gamma)}{ \max \left( N_e, 1 \right)}}\right\},
\end{equation}
where $|\mathcal{E}|$ is the total number of edges,  
and $\underline{\Psi} \geq 0$ is a known lower bound on all $\Psi_e,$ while $\gamma \in (0,1)$ is a parameter controlling the decay of expected regret\footnote{
In Supplementary Materials, we prove that under Alg-D the expected regret of a graph-restricted problem tends to $0$ as the number of episodes tends to $\infty$.}.
In each episode, we solve a shortest path problem based on edge weights $\hat{C}_e = - \log(1 - \hat{\Psi}_e)$. 
Then we simulate running through the derived optimal path and update edge statistics $N_e$ and $\phi_e$, which are used to change $\hat{C}_e$ in the next episode.
The resulting method is summarized in Algorithm \ref{alg:D}.

\begin{algorithm}
  \caption{Alg-D: model-based planning on a graph}
    \begin{algorithmic}\label{alg:D}
    \renewcommand{\algorithmicrequire}{\textbf{Input:}}
    \renewcommand{\algorithmicensure}{\textbf{Output:}}
    \STATE Set $\phi_e = 0, \; N_e = 0$ for all $e.$
      \WHILE {$t = 1:T$}
      	\STATE Update $\hat{\Psi}_e$ according to \eqref{eq:alg_D_ucb};	
	\STATE Solve the deterministic shortest path problem with edge costs $\hat{C}_e = - \log(1 - \hat{\Psi}_e )$;
	\STATE Simulate running through the $\hat{\Psi}$-optimal path from $\bv_0$ using the actual $\Psi$;
	\STATE Update $\phi_e$ and $N_e$ accordingly;
	\ENDWHILE
  \end{algorithmic} 
\end{algorithm}

Unfortunately, as we show in Section \ref{s:experiments}, the performance of such methods is rather poor in our continuous setting.
First, to ensure $\mathfrak{S}_j \rightarrow 0,$
the number of actions/edges per node would have to grow as we refine the graph -- otherwise, we will not be able to obtain all possible directions of motion in the limit.
The methods above are hard to use in MDPs with large action sets and are not directly usable in MDPs with infinite action spaces.
Second, in UCT any capture yields an equal penalty for all edges successfully traversed in that episode (even if the true $\Psi_e$ is quite low along some of them).  
This is why model-based methods, such as Alg-D,
are preferable for this class of problems even on graphs.  
Third, and most importantly, both UCT and Alg-D do not account for correlations in (unknown) transition functions of different (state, action) pairs.
In the continuous case, the smoothness of surveillance rate $K(\x)$ makes the spatial correlations crucial.  Ignoring this feature results in much slower learning.

In the next section, we overcome these limitations by introducing new methods for \emph{continuous model} learning and path planning based on a confidence-bounds-modified
version of the model.

\section{MODEL-BASED METHODS ON $\Omega$}
\label{s:ourAlgs}

\subsection{Piecewise-constant models of $K$ and planning based on confidence bounds}
\label{s:piecewise_constant}
% !TEX root = root.tex

As a first attempt to solve the continuous problem, we decompose $\Omega$ into a collection $\Gcell$ of non-overlapping subdomains/cells and assume that $K(\bx)$ is a constant on each of them.
We define three auxiliary objects 
to gather data inside each cell over many episodes:
\begin{itemize}
 \item $\Gcaught:$ the total number of captures
 in a cell;
 \item $\Gtime:$ the total time spent in a cell;
 \item $\Gvisit:$ the total number of visits/entries into a cell.
\end{itemize} 
We will use $\Ktilde \in R_{0,+}^{|\Gcell|}$ to represent a piecewise constant estimate of $K(\bx)$ and
$\sigtilde^2  \in R_{0,+}^{|\Gcell|}$ to denote the element-wise estimated variance of $\Ktilde$.

Focusing on a single cell, suppose that the surveillance intensity is indeed some (unknown) constant:  $K(\bx) = \Kcell.$
Enumerating all episodes in which our planned trajectory involved traveling through that cell and capture did not occur before we reached it, 
suppose in the $k$-th such episode our plan is to exit that cell after time $t_k$.
The capture time $S_k$ would be an exponentially distributed random variable with rate $\Kcell,$ but of course we
only get to observe its \textit{right censored} \citep{shorack2009empirical} version $R_k = \min(S_k, t_k)$.
For convenience, we also define a {\em capture indicator}  $\delta_k$, which is equal to $1$ if we are caught before exiting that cell and $0$ otherwise (implying $R_k = t_k$).
Assuming there were $n$ such visits to this cell up to the current episode,
this right-censored data $(\delta_k,R_k)_{k=1}^n$ can be used to derive the maximum likelihood estimate (MLE)
\begin{equation}\label{eq:mle}
 \Ktildecell = \sum_{k=1}^n \delta_k\Big/\sum_{k=1}^n R_k.
\end{equation}
The asymptotic expression for $\Ktildecell$'s variance is 
\begin{equation}
\label{eq:cell_estimate}
  \Kcell^2\Big/\sum_{k=1}^n [1 - \exp(- \Kcell t_k)].
\end{equation}
Using the fact that $\expect[\delta_k]=1 - \exp(- \Kcell t_k)$, we can estimate this asymptotic variance as
\begin{equation}\label{eq:mle_asymptotic_var}
  \sigtilde_{\text{cell}}^2 = \sum_{k=1}^n \delta_k \Biggm/ \left(\sum_{k=1}^n R_k\right)^2.
\end{equation}
Using $\Gcaught$ and $\Gtime$ notation, $\Ktilde$ and $\sigtilde$ can be written as $\Ktilde = \Gcaught/\Gtime, \sigtilde^2 = \Gcaught/\Gtime^2$ on each cell.

Inspired by the confidence bound techniques on graphs, we can also build up a ``lower-confidence'' intensity.
Since this modification can produce negative values and our 
observation intensity must be non-negative, we do not 
approximate $K$ directly but instead model $Z(\bx) = \log K(\bx).$
We use the statistic $\tilde{Z}_{\mathrm{cell}} = \log(\tilde{K}_{\mathrm{cell}})$
as an estimator for $Z$ values at a cell center, 
with $\tilde{K}_{\mathrm{cell}}$ defined as in \eqref{eq:cell_estimate}.
If $\tilde{K}_{\mathrm{cell}}$ is asymptotically distributed as
$N(\mu^{}_K, \sigma_K^2)$ with $\sigma_K^{}$ approaching 0, then by local
linearization of the logarithm (known as the {\em delta method}~\citep{van2000asymptotic}), we have that
$\tilde{Z}_{\mathrm{cell}}$ is asymptotically distributed as
$N(\log \mu^{}_K, \sigma_K^2 / \mu_K^2)$.  Using $\Gcaught$ and $\Gtime$ notation, we estimate the mean and variance for $\tilde{Z}_{\mathrm{cell}}$ by 
\begin{equation}
\label{eq:Z_estimate}
\tilde{Z} = \log\left(\mathcal{G}_c/\mathcal{G}_t\right), \quad \sigtilde_Z^2 = 1/\mathcal{G}_c.
\end{equation}
This allows us to define corresponding piecewise-constant functions $\tilde{Z}(\bx)$ and $\sigtilde_Z^{ }(\bx)$.
Our 
lower-confidence-adjusted intensity is constructed as
\begin{equation}\label{eq:ucb_K}
\Khat(\bx) = \exp\left( \tilde{Z}(\bx) - \sqrt{\log (T|\Gcell|/\gamma)} \sigtilde_Z^{ }(\bx) \right).
\end{equation}
Designed similarly to \eqref{eq:alg_D_ucb},  
the constant factor multiplying $\sigtilde_Z^{ }(\bx)$ 
balances the exploration vs exploitation and
depends on the total number of episodes $T$, the number of cells $|\Gcell|$, and $\gamma \in (0,1)$ controls the rate of exploration.
The resulting formula yields low values of $\Khat$ in rarely visited cells, thus encouraging the exploration if those cells are relevant for 
paths from $\bx_0$ to $\bdy$.  (E.g., a cell with a low $\Khat$ might be completely irrelevant if passing through it requires traversing other high-$\Khat$ cells
or if there exists a much shorter/safer path from $\bx_0$ to $\bdy$.) 
Since the above defined $\Khat$ is also piecewise-constant, the $\Khat$-optimal trajectory will be polygonal,
but finding its exact shape still requires solving an Eikonal equation.  We use a Fast Marching Method \citep{sethian1996fast} to do this on a finer grid  $\Gpredict$.

Algorithm \ref{alg:pw_const_ucb} summarizes the resulting method.
To avoid $\tilde{Z}$ in \eqref{eq:Z_estimate} becoming infinity, we initialize $\Gtime$ as a small positive constant $\epsilon$ and
$\Gcaught$ as $\epsilon K_{\text{init}}$ so that initially $\Gcaught/\Gtime$ equals to some constant $K_{\text{init}}$\footnote{
In our numerical tests, we chose $K_{\text{init}}$ to be the $\Omega$-averaged value of $K$.  We have also experimented with other values of $K_{\text{init}},$ but this did not seem to have significant effect on Algorithm's performance.}. 
\begin{algorithm}
  \caption{Alg-PC: planning with a piecewise-constant model}
    \begin{algorithmic}\label{alg:pw_const_ucb}
    \renewcommand{\algorithmicrequire}{\textbf{Input:}}
    \renewcommand{\algorithmicensure}{\textbf{Output:}}
    \REQUIRE $\gamma, \epsilon$.
    \STATE Set $\Gcaught = \epsilon K_{\text{init}},\, \Gtime = \epsilon$; 
    \STATE Set $\tilde{Z} = \log(\Gcaught/\Gtime),\, \sigtilde_Z^{ } = 1/\Gcaught$; 
      \WHILE {$t = 1:T$}
          \STATE $\Khat(\bx) = \exp\left( \tilde{Z}(\bx) - \sqrt{\log (T|\Gcell|/\gamma)} \sigtilde_Z^{ }(\bx) \right)$;
          \STATE Numerically solve $|\nabla \hat{u}(\bx)| = \Khat(\bx)$;
          \STATE Find an optimal path from $\bx_0$ using $\hat{u}(\bx)$;
          \STATE Simulate that path using the real $K(\x)$;
          \STATE Update $\Gcaught, \Gtime, \Gvisit$ using simulation results;
          \STATE Compute $\tilde{Z},\sigtilde_Z^{ }$ according to \eqref{eq:Z_estimate};
      \ENDWHILE
  \end{algorithmic} 
\end{algorithm}

\subsection{GP regression models of $K$ and planning based on confidence bounds}
\label{s:gaussian_process_regression}
% !TEX root = root.tex

Algorithm \ref{alg:pw_const_ucb} ignores the correlations between $K$
values in different cells and updates each cell independently.
In this section, we capture spatial correlation in the intensity by
using a Gaussian process
(GP)~\citep{williams2006gaussian}. A Gaussian
process is a collection of Gaussian random variables indexed by $\bx$,
with mean $m(\bx)$ and covariance $\Sigma(\bx, \bx')$.  
Some common choices of $\Sigma(\bx, \bx')$ are, e.g. the squared exponential kernel
\begin{equation}\label{eq:gp_kernel}
 \Sigma(\bx,\bx') = \alpha \exp\left( -|\bx - \bx'|^2/\beta^2 \right)
\end{equation}
or the Mat\'ern kernel 
\begin{equation}\label{eq:matern_kernel}
 \Sigma(\bx,\bx') = \alpha \frac{2^{1 - \nu}}{\Gamma(\nu)}\left( \sqrt{2\nu}d/\beta \right)^\nu B_\nu\left( \sqrt{2\nu}d/\beta \right)
\end{equation}
where $d = \left|\bx - \bx'\right|,$ $\Gamma$ is the gamma function, and $B_\nu$ is the modified Bessel function of the second kind.
Here, $(\alpha, \beta)$ are hyperparameters that can be learned from data while the parameter
$\nu$ in the Mat\'ern kernel controls the differentiability of GP and can reflect our assumptions about the level of regularity of $K(\bx).$

\textbf{Criteria$^*$}.
We only use cells whose estimates of $K$ are accurate enough as inputs of GP regression.
Here we introduce a list of rules to select these cells from $\Gcell$:
\begin{itemize}
 \item $\Gcaught \geq 1$, preventing
 $1/\Gcaught$
 from becoming infinity.
 \item $\Gvisit \geq n_{\min}$, guaranteeing enough entries into a cell.
 Our implementation uses $n_{\min}=20.$
 \item $\Gtime \geq t_{\min}$, avoiding extremely short traverses. 
  Our implementation sets  $t_{\min}$ to 
  the time sufficient to traverse a cell's diameter.
\end{itemize}

We denote the cells satisfying \textbf{Criteria}$^*$ as $\Gobserve \subset \Gcell$ and their centers as $\Xobserve$.
Let $\tilde{Z}_{\text{ob}},\sigtilde_{\text{ob}}$ be $\tilde{Z},\sigtilde^{}_Z$ values at $\Xobserve$ reshaped as vectors.
Use $\tilde{\Sigma}$ as an abbreviation of $\left[\Sigma_{\text{ob}} + \text{diag}(\sigtilde_{\text{ob}})\right]$.
The GP posterior mean $M(\bx)$ for $Z$ based on noisy observations at $\Xobserve$ is 
\begin{equation}\label{eq:gp_mean_log}
  M(\bx) = m(\bx) + \Sigma(\bx,\Xobserve)\tilde{\Sigma}^{-1}\big[\tilde{Z}_{\text{ob}} - m(\Xobserve)\big].
\end{equation}
Another advantage of GP model is that we simultaneously obtain the posterior covariance 
\begin{equation}\label{eq:gp_var_log}
  \rho^2(\bx) = \Sigma(\bx,\bx) - \Sigma(\bx,\Xobserve)\tilde{\Sigma}^{-1}\Sigma(\Xobserve,\bx).
\end{equation}
The resulting method is summarized in Algorithm \ref{alg:gp_log_ucb}.
We note that partial knowledge $K(\bx)$ can be encoded 
in a prior mean $m(\bx),$
though in our experiments $m(\bx)$ was simply set to a positive constant on the entire $\Omega.$
\begin{algorithm}
  \caption{Alg-GP: planning with a GP model}
    \begin{algorithmic}\label{alg:gp_log_ucb}
      \renewcommand{\algorithmicrequire}{\textbf{Input:}}
      \renewcommand{\algorithmicensure}{\textbf{Output:}}
      \REQUIRE $m(\bx), (\alpha, \beta), \gamma$ 
      \\ Set $\Gcaught,\Gtime,\Gvisit$ to be all zeros; 
      \\ Choose a kernel $\Sigma(\bx,\bx')$; 
      \\ $M(\bx) = m(\bx), \rho(\bx) = 0$;
      \WHILE {$t = 1:T$}
          \STATE $\Khat(\bx) = \exp(M(\bx) - \sqrt{\log(T|\Gcell|/\gamma)} \rho(\bx) )$;
          \STATE Numerically solve $|\nabla \hat{u}(\bx)| = \Khat(\bx)$;
          \STATE Find an optimal path from $\bx_0$ using $\hat{u}(\bx)$;
          \STATE Simulate that path using the real $K(\x)$;
          \STATE Update $\Gcaught, \Gtime, \Gvisit$ using simulation results;
          \STATE Compute $\tilde{Z},\sigtilde_Z^{ }$ according to \eqref{eq:Z_estimate};
          \STATE Determine $\Gobserve$ according to \textbf{Criteria}$^*$;
          \IF {$\Gobserve$ is non-empty}
          \STATE Update $M(\bx)$ and $\rho(\bx)$ using \eqref{eq:gp_mean_log} and \eqref{eq:gp_var_log};
          \ENDIF
          \IF {$t > 1000$ and $t \equiv 1$ mod(1000)}
          \STATE Tune hyperparameters $(\alpha, \beta)$ according to \eqref{eq:max_marginal_log_likelihood}.
          \ENDIF
      \ENDWHILE
  \end{algorithmic} 
\end{algorithm}

% !TEX root = root.tex

\subsection{Hyperparameter tuning}
The values of $(\alpha,\beta)$ are essential to the performance of GP regression.
GP provides a probabilistic framework for automatically selecting appropriate hyperparameters through maximizing $\log$ marginal likelihood \citep{2017-nips}: 
\begin{equation}\label{eq:max_marginal_log_likelihood}
  \max_{\alpha, \beta > 0}\ -\frac{1}{2}\z_{\text{ob,c}}^\intercal\tilde{\Sigma}^{-1}\z_{\text{ob,c}} - \frac{1}{2} \log |\tilde{\Sigma}| - \frac{n}{2} \log 2\pi,
\end{equation}
where $\z_{\text{ob,c}} = \tilde{Z}_{\text{ob}} - m(\Xobserve)$ is the vector
of centered observations.
In Algorithm \ref{alg:gp_log_ucb}, we conduct this hyperparameter tuning once every thousand episodes.

\section{NUMERICAL EXPERIMENTS}
\label{s:experiments}
% !TEX root = root.tex

We now compare the performance of 
Algorithms UCT, Alg-D, Alg-PC, and Alg-GP
on several examples\footnote{
In the interest of computational reproducibility, the source code of our implementation and additional experiments can be found at
\url{https://eikonal-equation.github.io/Bayesian-Surveillance-Evasion/}. 
GP updates, capture event simulations and the main loop of Alg-GP all Algorithms
are implemented in MATLAB while Fast Marching Method and the optimal path tracer are in C++.
We ran the experiments using Dell OptiPlex 7050 desktop with 3.6 GHz Intel i7-7700 processor, and 16 GB RAM.
With a $20\times 20$ observation grid, each of these examples took under 20 minutes with Alg-GP and under 5 minutes with Alg-PC.
} on the domain $\Omega = [0,1]^2.$
For simplicity, we assume $f(\bx) = 1$ and focus on 
different versions of $K(\bx)$, constructed as sums of several Gaussian peaks with different amplitudes and widths.
(Each peak might correspond to the location of a separate observation/surveillance center.)  
For each example, we conduct $T=15000$ episodes, always starting from the same initial position
$\src$ indicated by a cyan dot.

In all examples, we have benchmarked UCT on a graph $G$ built on a $20\times 20$ grid of nodes.  The same graph was also used to test the performance of Alg-D. 
In Alg-PC and Alg-GP, all examples use a $20\times 20$ observation grid $\Gcell$ 
while the Eikonal equations are solved on a $101\times 101$ grid $\Gpredict.$
In all cases, we have used $\lambda=\sqrt{2}$ and $\gamma = 0.1.$

Figures 1-3 provide detailed information on three representative examples, focusing on Alg-GPe (a version of Alg-GP using the squared exponential kernel).
In each case, the four subfigures (enumerated left to right, top to bottom) present:
(\romannumeral 1) The true surveillance intensity $K(\bx)$.
(\romannumeral 2) The level sets of $u(\bx)$; i.e., the minimum integral of $K$ along a trajectory starting at $\bx$.
(\romannumeral 3) The final $\exp(M(\bx))$; i.e.,  the Alg-GP-predicted $K(\bx)$ without confidence bound modification.
The magenta dots are locations of captures (only from experiments of Alg-GP).
The black curve is the optimal path based on $\exp(M(\bx))$ after the last episode.
(However, note that in each episode the trajectory is planned according to a version of $\Khat(\bx)$ available at that time.)
(\romannumeral 4) The final GP posterior variance $\rho(\bx)$.
Figure 4 shows the regret metric (i.e., the excess rate of captures $\mathfrak{S}$) for all of the benchmarked algorithms.

\begin{figure}[h]
 \centering
 \includegraphics[width=0.95\linewidth]{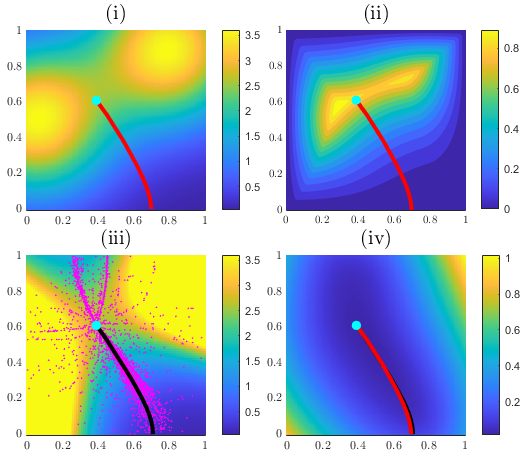}
\caption{\label{fig:double_observer}
Bimodal surveillance intensity $K(\bx)$.
Most of the selected paths are around three locally optimal paths with the longer one being globally optimal.
}
\end{figure}
\begin{figure}[h]
 \centering
 \includegraphics[width=0.95\linewidth]{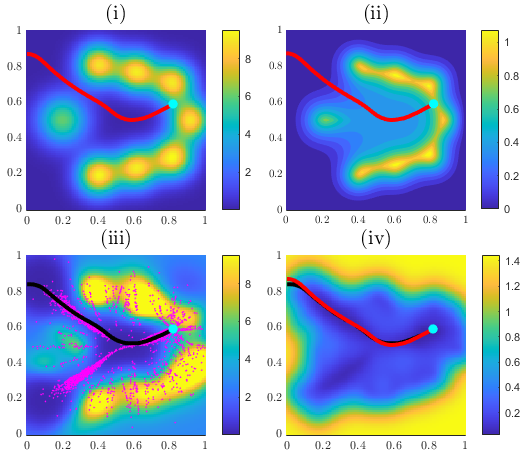}
\caption{\label{fig:multi_observer}
Surveillance intensity inspired by Figure 7 from \citep{cartee2020control}. 
The shortest exiting path induces a higher capture probability, while the actual optimal path takes a longer detour towards a lower intensity region.
Selected paths mostly cluster around two locally optimal paths, with one of them being globally optimal.
}
\end{figure}
\begin{figure}[h]
 \centering
 \includegraphics[width=0.95\linewidth]{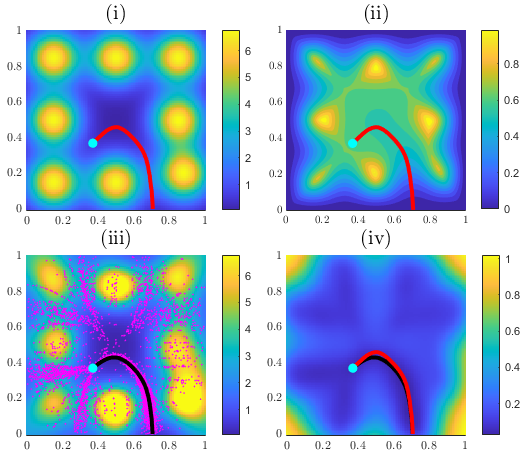}
\caption{\label{fig:misplaced_observer}
An intensity with eight peaks and multiple locally optimal paths.  
The peak around the southeast corner is displaced slightly, creating a gap in $K(\bx)$ and the optimal path reflects this.
Alg-GP attempts to discover each locally optimal path but concentrates more around three paths with one of them being globally optimal.
}
\end{figure}
\begin{figure}[hhhh]
 \centering
 \includegraphics[width=0.98\linewidth]{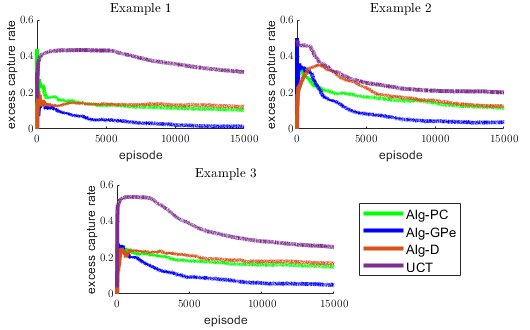}
\caption{\label{fig:example_regrets}
Regret metric (excess capture rate $\mathfrak{S}$) for examples illustrated in Figures  \ref{fig:double_observer}, \ref{fig:multi_observer} and \ref{fig:misplaced_observer}.
}
\end{figure}
The following observation hold true for all examples:\\
$\bullet \;$ Capture locations (magenta dots) in subfigure (\romannumeral 3) give rough indications of which parts of $\Omega$ Alg-GP prefers to explore.
It mostly selects paths around locally optimal ones, and during the later episodes it focuses more on the vicinity of the globally optimal one, showing a proper balance between exploration and exploitation.
Indeed, the two paths in subfigure (\romannumeral 4) indicate that Alg-GP's final prediction (black) approximately matches the true optimal path (red).\\
$\bullet \;$ In Figure \ref{fig:example_regrets}, Alg-GP's averaged excess capture rate continuously decreases and appears to confirm the convergence.
The third example contains multiple locally optimal (and roughly comparable) paths and requires more episodes to learn the globally optimal one, which explains the slow decrease of $\mathfrak{S}$ in later episodes.\\
$\bullet \;$ In contrast, Alg-D and Alg-PC exhibit a consistently slower improvement of $\mathfrak{S}$ and even stagnation -- indicating that model-learning takes substantially longer for these methods.\\
$\bullet \;$ UCT  algorithm generates much larger regrets than all others.
We have also tested it with more episodes ($10^5 \sim10^6$), but the results still show obvious stagnation.
The main reason is that UCT learns the state-action functions directly.
It regards a whole path as a single datum and overlooks the information of not being captured along the earlier parts of the path, which could be used to improve the estimates for $K$.\\
$\bullet \;$ Both Alg-D and UCT restrict the Evader to move only along one of the eight directions.
In principle, this will result in a gap between the truly optimal $W_*$ and the risk along the best path on this graph even after infinitely many episodes.
However, as we show in Supplementary Materials, in these examples that gap is much smaller than the observed regret $\mathfrak{S}$.

{\bf Remarks on grid effects:}  
Our use of a discrete observation grid $\Gcell$ essentially lumps together all captures within one cell
and limits the algorithm's ability to learn the true $K$.  To account for this more accurately, we could
modify our definition of the regret metric,
measuring the regret relative to the best path learnable on the specific $\Gcell.$
We include such detailed tests in 
Supplementary Materials.
But in summary we note that they
demonstrate yet another advantage of GP regression:

1) 
For Alg-GP, if we assume infinitely many captures in each cell,
the ideal (asymptotically learnable) $\tilde{W}_* = 1 - \exp(\J_*) $ is only very weakly dependent on $\Gcell.$
E.g.,  the $\tilde{W}_*$ is already within 0.25\% from the truly optimal
$W_* = 1 - \exp( - u(\src))$  for all 
examples considered above even with a coarse $10\times 10$ observation grid.

2)
For Alg-PC, the ideal learnable $\tilde{W}_*$ can improve significantly when we refine $\Gcell.$
Asymptotically, the averaged regret is certainly better for finer observation grids.
But they also present a challenge since many more episodes are needed to obtain a reasonable approximation of $\tilde{K}$
in all potentially relevant cells before the finer grid's asymptotic advantage becomes relevant.  E.g., re-running the example from Figure~\ref{fig:multi_observer} with $T=60,000,$ 
Alg-PC yields a lower averaged regret on a  $10\times 10$ observation grid than on a  $40\times 40$ grid for the first
50,000 episodes, while the averaged regret on a $20\times 20$ grid is much smaller throughout.

3)
The effects of the computational grid $\Gpredict$ appear to be negligible in all of our simulations. The errors due to discretizing the Eikonal PDE are dominated by 
the errors due to uncertainty in $K$ and the $\Gcell$ effects described above.

{\bf Computational complexity:}
The cost of each episode consists of two components: (a) updating surveillance intensity model and (b) HJB-based path-planning using the current model.  The cost of (a) for Alg-PC scales as $O(|\Gcell|);$ i.e., linearly in the number of subdomains/cells.
For Alg-GP,
the complexity of this stage depends on the size of $\mathcal{G}_{ob}$ (the collection of cells satisfying \textbf{Criteria$^*$}).
During each episode, the cost of Alg-GP's update is dominated by solving the linear system in updating posterior covariance \eqref{eq:gp_var_log}, which is $O(|\mathcal{G}_{ob}|^3)$ and $|\mathcal{G}_{ob}|$ is usually quite small compared to $|\Gcell|$.
Alg-GP also includes an additional cost of periodically re-tuning the hyperparameters.

The use of the Fast Marching Method makes the actual path planning in each episode quite fast even for much finer $\Gpredict.$
The cost of (b) is dominated by solving the PDE on a grid, which in our setting is $O(N \log N)$ on a discretization grid $\Gpredict$ with $N$ gridpoints.  The latter cost can be prohibitive in high-dimensional generalizations since $N$ grows exponentially with the dimension of $\Omega.$  The usual approach to overcome this ``curse of dimensionality'' is to use Approximate DP (e.g., based on mesh-free HJB discretizations), but this is not necessary in our 2D setting.

\begin{figure}[h]
 \centering
 \includegraphics[width=0.98\linewidth]{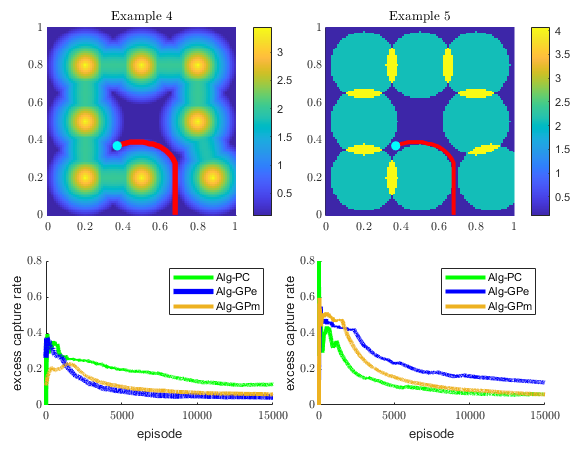}
\caption{\label{fig:nonsmooth_K}
Two examples of non-smooth surveillance intensity. 
Example 4 (LEFT): a non-smooth-yet-continuous $K(\bx)$. 
Example 5 (RIGHT): a discontinuous $K(\bx)$. 
The Mat\'ern kernel with $\nu = 1/2$, which assumes less smoothness of the interpolated function, performs better than the squared exponential kernel.
}
\end{figure}

{\bf Learning less regular $K$:}  
We have also used Examples 1-3 to test Alg-GP with the Mat\'ern kernel, taking $\nu = 5/2$.
The results are quite similar to those of Alg-GPe and thus omitted.  But the situation is noticeably different when the actual $K(\x)$ is non-smooth or even discontinuous.
Figure \ref{fig:nonsmooth_K} shows two such examples, modifying the smooth intensity from Figure \ref{fig:misplaced_observer}.
Suppose the center of a peak is $\bx_c$, our first example replaces each single Gaussian peak by
$K_{\text{single}}(\bx) = \max\{0,\, K_p - |\bx - \bx_c|\}$ where $K_p > 0$ is the maximum.
Such a surveillance intensity (used in Example 4) is still continuous but not smooth.
In Example 5, we replace each Gaussian peak by a piecewise constant function: $K_{\text{single}}(\bx) = K_c$ when $|\bx - \bx_c| \leq r$ for some radius $r > 0$ while $K_{\text{single}}(\bx) = 0$ otherwise.
In both cases, a small positive constant was later added to ensure that the resulting $K(\bx) > 0$ for all $\x \in \Omega.$

We have conducted these experiments with both the squared exponential kernel (Alg-GPe) and Mat\'ern kernel (Alg-GPm) with $\nu = 1/2$.
In the continuous but non-smooth Example 4, Alg-GPe still remains the best though Alg-GPm is 
almost as good.
In the discontinuous Example 5, Alg-PC actually initially outperforms both GP-based algorithms though Alg-GPm eventually catches up.

\section{CONCLUSIONS}
\label{s:conclusions}
% !TEX root = root.tex

We developed and numerically tested three algorithms (Alg-D, Alg-PC, and Alg-GP) for continuous path-planning problems with unknown random termination/capture intensity.
These algorithms follow a Bayesian approach
to model the surveillance intensity $K$ 
and then apply confidence bound techniques to tackle the exploration-exploitation dilemma.
The GP-regression 
used in Alg-GP leverages the spatial correlations in $K$ and usually results in more efficient learning from
captures
-- particularly when the GP covariance kernel is chosen to reflect the smoothness of the actual $K$.
While our experimental results are very promising, we do not currently have a proof of convergence and rigorous upper bound on the cumulative regret.
We hope that the graph-theoretic UCB proofs \citep{azar2017minimax} can be extended to cover edge-correlations and ultimately to our continuous setting.

Several extensions would broaden the applicability of our approach. 
If we consider E's post-detection planning, the terminal cost will become spatially inhomogeneous \citep{andrews2014deterministic}.
Multiobjective control methods \citep{kumar2010efficient} and various robust path planning techniques \citep{qi2021optimality} will become relevant
if the risk of detection is balanced against other optimization criteria (e.g., the profit from smuggling resources from a protected area 
\citep{Arnold_2019, cartee2020control, chen2021modeling}).
We have focused on the isotropic dynamics primarily to simplify the exposition.
Non-isotropic controlled dynamics lead to more general HJB PDEs, which can be similarly solved on a grid to approximate the value function efficiently \citep{sethian2001ordered, TsaiChengOsherZhao, AltonMitchell2, Mirebeau3}.  The gradient of that value function can be then used to synthesize the optimal control.
If $K$ is viewed as changing in time, this will introduce an additional challenge of change point detection \citep{aminikhanghahi2017survey} in our RL algorithms.
It will be also interesting to consider an antagonistic version, where $K$ is chosen by surveillance authorities to maximize the probability of capture in response to E's path choices.
A Nash equilibrium for this problem was already studied under the assumption that $K$ is selected (perhaps probabilistically) from a finite 
list of options $K_1,...,K_r,$ all of which are known to E \citep{gilles2020evasive, cartee2019time}.  We hope that it can be also extended to our setting where $K_i$'s are learned online.

\subsubsection*{Acknowledgements}
This research was supported in part by the NSF DMS (awards 1645643, 1738010, and 2111522). 
The authors are also grateful to M.~Wegkamp and M.~Nussbaum for advice on handling the right-censored data.

\bibliography{mybibfile}

%%%%%%%%%%%%%%%%%%%%%%%%%%%%%%%%%%%
%%%%%% SUPPLEMENT (OPTIONAL) %%%%%%
%%%%%%%%%%%%%%%%%%%%%%%%%%%%%%%%%%%
\appendix
% !TEX root = root.tex
% For one-column format, uncomment the following:
\onecolumn
% For two-column format, uncomment the following:
%\twocolumn[ \makesupplementtitle ]
\aistatstitle{Supplementary Materials}

\section{PROOF OF UPPER BOUND FOR Alg-D}

In this section we modify the upper bound proof of UCB-VI Algorithm \citep{agarwal2019reinforcement} to develop similar upper bounds on expected regret for our Alg-D described in section \ref{s:discrete_ucb_algorithm}.
We use superscript $t$ to indicate the corresponding episode that a quantity belongs to.
Recall that $\Psi_e$ denotes the true probability of capture when attempting to traverse an edge $e \in \mathcal{E}$ while $\tilde{\Psi}^t_e = \phi_e^t / N^t_e$ denotes our current best estimate for $\Psi_e$ based on the attempted traversals of $e$ and captures on $e$ up to the episode $t$.

We assume that there exist constants $\underline{\Psi}$ and $\overline{\Psi}$ such that $0 < \underline{\Psi} \leq \Psi_e \leq \overline{\Psi} < 1$ and $-\log(1 - x)$ is $L$-Lipschitz on $[\underline{\Psi}, \overline{\Psi}]$.
Since $C_e = -\log(1 - \Psi_e)$ and $\tilde{C}_e^t = -\log(1 - \tilde{\Psi}_e^t)$, we know that
\[
\prob\big(|\tilde{C}_e^t - C_e| \geq L\epsilon\big) \leq  \prob\big(|\tilde{\Psi}_e^t - \Psi_e| \geq \epsilon\big).
\]
For the estimate $\tilde{\Psi}_e^t$, Hoeffding's inequality leads to 
\[
\prob\big(|\tilde{\Psi}_e^t - \Psi_e| \geq \epsilon\big) \leq 2\exp(-2N_e^t\epsilon^2),
\]
where $N_e^t$ is how many times edge $e$ is visited up until the $t$-th episode.
Choosing $\epsilon = \sqrt{\frac{\log(T |\mathcal{E}|/\gamma)}{N_e^t}}$, we obtain
\[
 \prob\left(\left|\tilde{\Psi}_e^t - \Psi_e\right| \geq \sqrt{\frac{\log(T |\mathcal{E}|/\gamma)}{N_e^t}}\right) \leq \frac{2\gamma^2}{T^2|\mathcal{E}|^2}.
\]
Applying Boole's inequality over all edges and all episodes, we obtain a union bound  
\[
 \prob\left(\bigcup\limits_{e\in \mathcal{E}, t\leq T}\left|\tilde{C}_e^t - C_e\right| \geq L\sqrt{\frac{\log(T |\mathcal{E}|/\gamma)}{N_e^t}}\right) \leq \sum\limits_{e\in \mathcal{E}, t\leq T}\prob\left(\left|\tilde{C}_e^t - C_e\right| \geq L\sqrt{\frac{\log(T |\mathcal{E}|/\gamma)}{N_e^t}}\right)
\]
\[
 \leq \sum\limits_{e\in \mathcal{E}, t\leq T}\prob\left(\left|\tilde{\Psi}_e^t - \Psi_e\right| \geq \sqrt{\frac{\log(T |\mathcal{E}|/\gamma)}{N_e^t}}\right)  \leq \frac{2\gamma^2}{T|\mathcal{E}|}.
\]

This equation measures the model error since it bounds the difference between the estimated cost with the true cost.
With probability at least  $(1 - 2\gamma^2/T|\mathcal{E}|)$, $|\tilde{C}_e^t - C_e| \leq L\sqrt{\frac{\log(T |\mathcal{E}|/\gamma)}{N_e^t}}$ for any edge and any episode. 
Along the $t$-th episode's path $p^t$, triangle inequality yields
\[
 \left|\sum_{e\in p^t} \tilde{C}_e^t
  - \sum_{e\in p^t} C_e\right| \leq \sum_{e\in p^t} \left|\tilde{C}_e^t - C_e\right| \leq L\sqrt{\log(T|\mathcal{E}|/\gamma)}\sum_{e\in p^t}\frac{1}{\sqrt{N_e^t}}. 
\]
We denote the last term as $\Delta_{p^t}$. 
As $\left| \hat{\Psi}_e^t - \tilde{\Psi}_e^t\right| \leq \sqrt{\frac{\log(T |\mathcal{E}| / 
 \gamma)}{N_e^t}}$, using the Lipschitz condition of $-\log(1 - x)$ we obtain
\[
 \left|\hat{C}_e^t - \tilde{C}_e\right| \leq L\sqrt{\frac{\log(T |\mathcal{E}| / 
 \gamma)}{N_e^t}}.
\]
Therefore, by triangle inequality we have
\[
 \left|\sum_{e\in p^t} \hat{C}_e^t
  - \sum_{e\in p^t} C_e\right| \leq \left|\sum_{e\in p^t} \hat{C}_e^t
  - \sum_{e\in p^t} \tilde{C}_e^t \right| + \left|\sum_{e\in p^t} \tilde{C}_e^t
  - \sum_{e\in p^t} C_e\right| \leq 2\Delta_{p^t}. 
\]
We denote $v_{p^t}$ the true total cost of $p^t$ (i.e., the sum of true edge costs $C_e$ along $p^t$) and $\hat{v}_{p^t}$ the total cost based on the modified cost $\hat{C}_e.$
Using $v_{p^t}= \sum_{e\in p^t} C_e$ and $\hat{v}_{p^t} = \sum_{e\in p^t} \hat{C}_e^t$, we finally obtain
\[
 \prob\big(\forall t, \left|\hat{v}_{p^t} - v_{p^t}\right| \leq 2\Delta_{p^t} \big) \geq 1 - \frac{2\gamma^2}{T|\mathcal{E}|}.
\]

Notice that either $\hat{C}_e^t = -\log(1 - \underline{\Psi}) \leq C_e$ or $\hat{\Psi}_e^t = \tilde{\Psi}_e^t - \sqrt{\frac{\log(T|\mathcal{E}|/\gamma)}{N_e^t}} \leq \Psi_e$ (hence $\hat{C}_e^t \leq C_e$) with the probability of at least  $q = (1 - \frac{2\gamma^2}{T|\mathcal{E}|})$. 
Both cases lead to 
\[
 v_{p^t} -2 \Delta_{p^t} \leq \hat{v}_{p^t} \leq \hat{v}_{p^*} = \sum_{e\in p^*} \hat{C}_e^t \leq \sum_{e\in p^*} C_e = v_{p^*},
\]
where the second inequality follows from the $\hat{C}_e^t$-optimality of $p^t$ while the first \& third inequalities hold with probability of at least  $q.$

Assuming that each path the algorithm considers has no more than $M \leq |\mathcal{E}|$ edges, the expected cumulative regret can be bounded as
\[
 \expect\left[\sum_{t=1}^T (v_{p^t} - v_{p^*}) \right] \leq \prob\big(\forall t,\, v_{p^t} - v_{p^*} \leq 2\Delta_{p^t} \big)\sum_{t=1}^T\left(v_{p^t} - v_{p^*}\right) +\, \prob\big(\exists t,\, v_{p^t} - v_{p^*} > 2\Delta_{p^t} \big) \sum_{t=1}^T M \overline{q}
\]
\[
 \leq 2L\sqrt{\log(T|\mathcal{E}|/\gamma)} \sum_{t=1}^T\sum_{e\in p^t}\frac{1}{\sqrt{N_e^t}} + \frac{2\gamma^2 M\overline{C}}{|\mathcal{E}|},
\]
where $\overline{C} = -\log(1 - \overline{\Psi})$.

\begin{lemma}
\label{lemma:sum_freq}
The first term in previous inequality satisfies
\[
 \sum_{t=1}^T\sum_{e\in p^t}\frac{1}{\sqrt{N_e^t}} \leq \sqrt{MT|\mathcal{E}|}.
\]

\begin{proof}
A different way to do the summation leads to
\[
  \sum_{t=1}^T\sum_{e\in p^t}\frac{1}{\sqrt{N_e^t}} = \sum_{e\in \mathcal{E}}\sum_{i=1}^{N_e^T} \frac{1}{\sqrt{i}} \leq \sum_{e\in \mathcal{E}} \sqrt{N_e^T} \leq \sqrt{|\mathcal{E}|}\sqrt{\sum_{e\in \mathcal{E}} N_e^T} \leq \sqrt{MT|\mathcal{E}|}.
\]
The first inequality is due to the Jensen's inequality while the second follows from the Cauchy-Schwartz inequality.
The last one uses the fact that there are no more than $MT$ visits to all the edges.
\end{proof}
\end{lemma}

\begin{theorem}
\label{thm:regret_frequency}
Using Lemma \ref{lemma:sum_freq}, we can bound the averaged expected regret of our Alg-D on graph as 
\begin{equation}
\label{eq:regret_bound}
\frac{1}{T}\expect\left[\sum_{t=1}^T (v_{p^t} - v_{p^*}) \right] \leq \frac{2L}{T}\sqrt{2MT|\mathcal{E}|\log(T|\mathcal{E}|/\gamma)} + \frac{2\gamma^2 M\overline{C}}{T|\mathcal{E}|}.
\end{equation}
In particular, as $T\rightarrow \infty$, the averaged expected regret converges to zero.
\end{theorem}
We note that the above bounds apply to a graph-constrained problem only.  The graph-optimal capture probability $\left( 1- \exp[-v_{p^*}] \right)$ might in principle be significantly larger than the $W_*= 1 - \exp[ - u(\src)]$ describing the minimized capture probability in the continuous domain $\Omega.$

\section{HYPERPARAMETER TUNING}

We conduct hyperparameter tuning \eqref{eq:max_marginal_log_likelihood} every thousand episodes by applying the constrained optimizer \texttt{fmincon} with gradients in MATLAB .
Assume the log marginal likelihood in \eqref{eq:max_marginal_log_likelihood} is $\log\left(\tilde{Z}_{\text{ob}} \Big|X_{\text{ob}}, \alpha, \beta \right)$, the $\alpha$-gradient can be computed as
\begin{equation}
\frac{\partial}{\partial \alpha} \log\left(\tilde{Z}_{\text{ob}} \Big|X_{\text{ob}}, \alpha, \beta \right) = \frac{1}{2} \z_{\text{ob,c}}^\intercal\tilde{\Sigma}^{-1}\frac{\partial \tilde{\Sigma}}{\partial \alpha} \tilde{\Sigma}^{-1} \z_{\text{ob,c}} -\frac{1}{2} \text{tr}\left( \tilde{\Sigma}^{-1} \frac{\partial \tilde{\Sigma}}{\partial \alpha} \right).
\end{equation}
$\z_{\text{ob,c}}^\intercal$ and $\tilde{\Sigma}$ are the same notations as defined in section \ref{s:gaussian_process_regression}.
The expression of $\frac{\partial \tilde{\Sigma}}{\partial \alpha}$ depends on which kernel we choose.
The $\beta$-gradient can be computed similarly.

\section{AVERAGED EXCESS RISK $\mathfrak{R}$}
In this section, we define another path-related regret metric in addition to the excess rate of captures defined in \eqref{eq:regret_measures_2}.
If $\y_i(\cdot)$ is the $i$th-episode path and $\J_i$ is the corresponding cumulative intensity, the capture probability along $\y_i(\cdot)$ is $W_i = 1 - \exp(-\J_i)$.
The minimum capture probability is $W_* = 1 - \exp( - u(\src))$.
The \emph{averaged excess risk} is defined as
\begin{equation}\label{eq:averaged_excess_risk}
  \mathfrak{R}_j = \frac{1}{j}\sum_{i=1}^j (W_i - W_*), \ j = 1,2,\cdots,T.
\end{equation}
Unlike excess rate of captures which uses probabilistic outcomes, $\mathfrak{R}$ compares directly  the expected capture rate of episodic paths with the truly optimal $W_*$. 
Unfortunately, UCT algorithm is not guaranteed to find a path reaching the boundary during each episode.
As a result, $W_i$ cannot be computed and regret $\mathfrak{R}$ is not applicable to UCT. 
The following are $\mathfrak{R}$ results of Alg-D, Alg-PC and Alg-GP for all examples in this paper.
\begin{figure}[H]
\centering
 \includegraphics[width=0.95\linewidth]{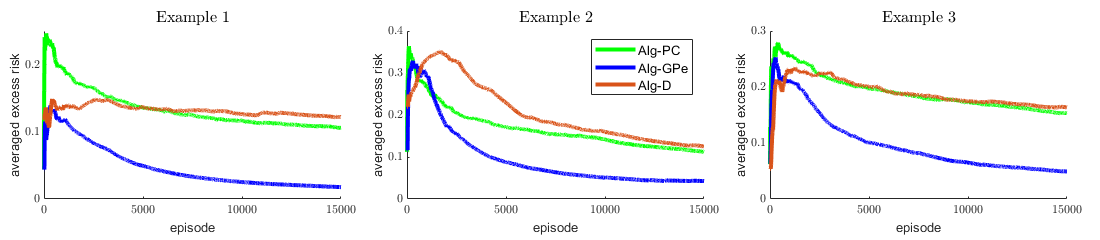}
\caption{\label{fig:average_excess_risk}
Regret metric (averaged excess risk $\mathfrak{R}$) for examples illustrated in Figures  \ref{fig:double_observer}, \ref{fig:multi_observer} and \ref{fig:misplaced_observer}.
}
\end{figure}
\begin{figure}[h]
\centering
 \includegraphics[width=0.7\linewidth]{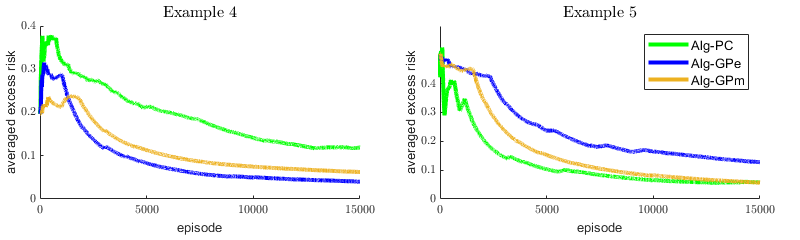}
\caption{\label{fig:average_excess_risk_nonsmooth}
Regret metric (averaged excess risk $\mathfrak{R}$) for examples illustrated in Figures \ref{fig:nonsmooth_K}.
}
\end{figure}

\section{LEARNABLE LIMITS RESTRICTED BY DISCRETIZATIONS}
\subsection{UCT and Alg-D}
For graph algorithms UCT and Alg-D, the Evader selects paths only on the grid and is restricted to move only along eight fixed directions.
As a result, even if we let the grid size approach zero, in the limit there is a gap between $W_*$ and the minimum capture rate that can be achieved on the grid.
From the figure below we can observe, as the grid size decreases, the blue dots approach some level which is closer to the optimal (red line, original continuous case, computed on a $h = 1/800$ grid) but there remains a gap ($\approx 0.014$ when $h = 1/640$).
Our numerical experiments use $h = 1/20$, in which case the best learnable limit is larger than the optimal capture probability by about 0.033.
However, as we observe from Figure \ref{fig:multi_observer}, after 15000 episodes the regret of Alg-D remains much larger than 0.033; i.e. the error due to this grid-restricted motion is not the reason why Alg-D shows such huge regrets.
Using a smaller grid size can reduce this gap, but it also increases the number of parameters to be estimated.
An overly fine grid takes more episodes to obtain accurate enough prediction of the capture rate, making it harder to observe its advantages.
\begin{figure}[h]
\centering
 \includegraphics[width=0.75\linewidth]{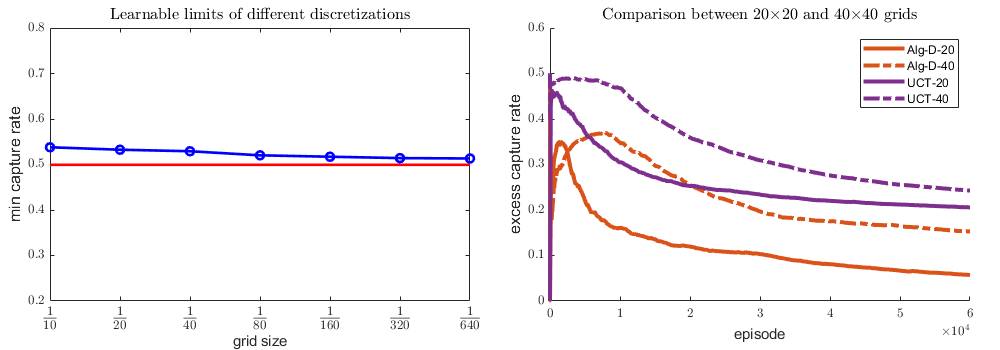}
\caption{\label{fig:graph_learnable_limits}
Left: the minimum capture probability over different discretized grids for the example in Figure \ref{fig:multi_observer}.
Right: We run UCT and Alg-D on 40 $\times$ 40 grids and compare regret $\mathfrak{S}$ with the results of 20 $\times$ 20 grids for 60000 episodes.
We can observe from the figure that using a 20 $\times$ 20 grids generates much smaller regrets than 40 $\times$ 40.
}
\end{figure}

\subsection{Alg-PC and Alg-GP}
We conduct an observation grid refinement study for Alg-PC using Example 2. 
The following figure shows the $\log$ of (non-averaged, instantaneous) differences between $W_i$ and $W_*$ in each episode, for different $\Gcell$ resolutions.
We observe that the lower $(W_i - W_*)$ values for $20\times 20$ and $40\times 40$ grids are much lower than for a $10 \times 10$ grid.
This indicates that with a finer grid Alg-PC is able to explore paths which are closer to optimal.
\begin{figure}[H]
\centering
 \includegraphics[width=0.4\linewidth]{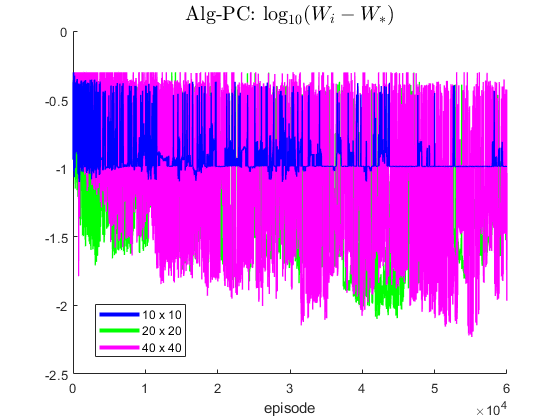}
\caption{\label{fig:algpc_refine}
Instantaneous differences between $W_i$ and $W_*$ in each episode. The grids are $10\times10, 20\times 20$ and $40\times 40$.
}
\end{figure}
Our use of the discrete observational grid $\Gcell$ essentially lumps together all captures within one cell.
This limits the algorithm's ability to learn the true $K.$
To account for this more accurately, we can modify our definition of regret metric and measure the regret relative to the best path learnable on the specific $\Gcell$.  
Below we provide additional tests illustrating this idea for Alg-PC and data showing that this subtlety is largely irrelevant for Alg-GP.

We define $\tilde{K}_*(\bx)$ to be the cell-averaged version of $K.$
This piecewise-constant function also represents the best approximation of $K$ that we could hope to attain with infinitely many captures in every cell.
Computing the viscosity solution to $| \nabla \tilde{u}_* | = \tilde{K}_*,$ we obtain the $\Gcell$-optimal feedback control $\tilde{\ba}_* = -\nabla \tilde{u}_* /|\nabla \tilde{u}_*|$ and can compute its actual quality $w(\bx)$ by integrating the true $K$ over the paths resulting from $\tilde{\ba}_*$.
The corresponding capture probability is then $\tilde{W}_* = 1 - \exp( -w(\src))$ and we can define the $\mathcal{G}$-adjusted regret by using $\tilde{W}_*$ instead.
\begin{figure}[H]
\centering
 \includegraphics[width=0.9\linewidth]{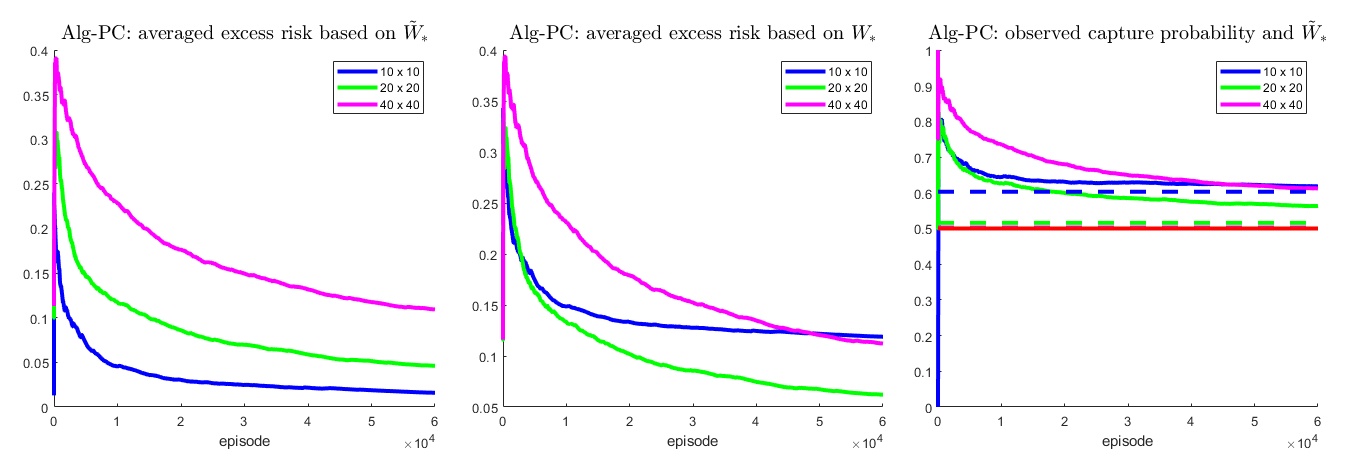}
\caption{\label{fig:algpc_learnable_limits}
Left: the new regret defined as above for different sizes of $\Gcell$, applying Alg-PC to Example 2.
It illustrates that finer $\Gcell$ requires more episodes to approach the asymptotic learnable limit.
But even though the $20 \times 20$ $\Gcell$ results in a higher grid-adjusted regret than the $10 \times 10$ grid, this is not the case for the regret-against-the-truly-optimal-$W_*$, as illustrated in the second (center) figure.
The third (right) figure illustrates the experimentally observed capture rate (note that this is not the regret metric $\mathfrak{S}$), together with $\tilde{W}_*$ (dashed lines) for different sizes of $\Gcell$.
We observe that $\tilde{W}_*$ becomes closer to the true optimal $W_*$ (the red horizontal line) as $\Gcell$ becomes finer.
But it takes about $50,000$ episodes until the asymptotic advantage of the $40 \times 40$ grid yields a lower capture rate than on the $10 \times 10$.
In this example, the $20 \times 20$ grid yields much smaller regrets than the other two grids after $10,000$ episodes.
}
\end{figure}

Gaussian process regression performs surprisingly well in recovering the optimal path, even on a coarse grid.
For all examples considered in the paper, $\tilde{W}_*$ is already within 0.25\% from the truly optimal $W_*$ even with a coarse $10\times 10$ observation grid.
The following is an illustration of this phenomenon using the second and third examples from the paper.
\begin{figure}[H]
\centering
 \includegraphics[width=0.88\linewidth]{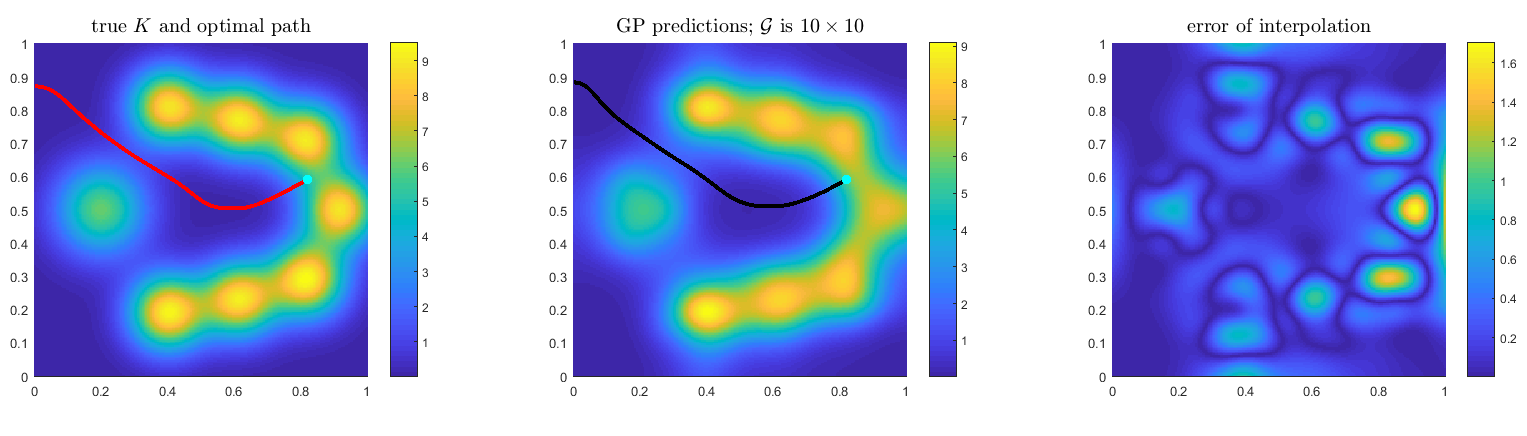}
 \includegraphics[width=0.88\linewidth]{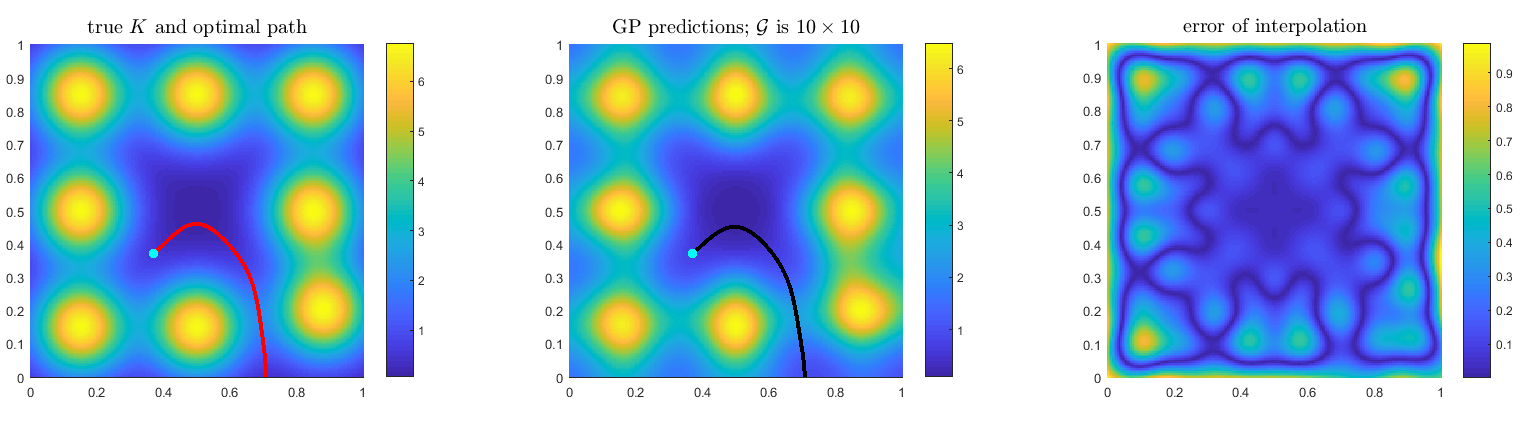}
\caption{\label{fig:gp_approximation_power}
Left: true $K(\bx)$ and the truly optimal path. Center: GP-predicted intensity with an infinite number of observations on a $10 \times 10$ grid $\Gcell$ and the corresponding ``optimal'' path. Right: interpolation errors. 
Top row: example 2; bottom row: example 3.  In both cases, the GP-predicted optimal path is quite close to the true optimal path even though the GP-estimated $\tilde{K}_*$ is quite different from the true $K(\bx).$
}
\end{figure}

\end{document}